\documentclass[letterpaper]{article} 
\usepackage{aaai24}  
\usepackage{times}  
\usepackage{helvet}  
\usepackage{courier}  
\usepackage[hyphens]{url}  
\usepackage{graphicx} 
\usepackage{natbib}  
\usepackage{caption} 
\usepackage{array}
\usepackage{amsmath}
\usepackage{algorithm}
\usepackage{algorithmic}

\usepackage{newfloat}
\usepackage{listings}
\DeclareCaptionStyle{ruled}{labelfont=normalfont,labelsep=colon,strut=off} 
\floatstyle{ruled}
\newfloat{listing}{tb}{lst}{}
\floatname{listing}{Listing}
\title{Validation, Robustness, and Accuracy of Perturbation-Based Sensitivity Analysis Methods for Time-Series Deep Learning Models}

\title{Validation, Robustness, and Accuracy of Perturbation-based Sensitivity Analysis Methods for Time-Series Deep Learning Models}
\author {
    Zhengguang Wang \\
    (Advisor: Judy Fox)
}
\affiliations{
    University of Virginia\\
    Computer Science Department \\
    zw4re@virginia.edu \\
}

\begin{document}

\maketitle

\begin{abstract}
This work undertakes studies to evaluate Interpretability Methods for Time-Series Deep Learning. Sensitivity analysis assesses how input changes affect the output, constituting a key component of interpretation. Among the post-hoc interpretation methods such as back-propagation, perturbation, and approximation, my work will investigate perturbation-based sensitivity Analysis methods on modern Transformer models to benchmark their performances. Specifically, my work intends to answer three research questions: 1) Do different sensitivity analysis methods yield comparable outputs and attribute importance rankings? 2) Using the same sensitivity analysis method, do different Deep Learning models impact the output of the sensitivity analysis? 3) How well do the results from sensitivity analysis methods align with the ground truth?
\end{abstract}

\section{Introduction}

Deep Learning techniques for analyzing time-series data and developing associated interpretation methods have gained significant importance, given that a substantial portion of the world's information is inherently time-dependent.
With state-of-the-art deep learning models and appropriate interpretation methods, we identify temporal patterns from the time-series data. The outcomes can be used for prediction and inference tasks in many fields, thus aiding scientific discoveries and helping governments make decisions. 

I want to undertake studies to evaluate \textbf{Interpretability Methods for Time-Series Deep Learning}. Sensitivity analysis assesses how input changes affect the output, constituting a key component of interpretation. Among the post-hoc interpretation methods such as back-propagation, perturbation, and approximation, I will investigate perturbation-based sensitivity Analysis methods on modern Transformer model to benchmark their performances.
My research focuses on evaluating the validation, robustness, and accuracy of these methods and the social impact of their applications. For example, if these methods yield a similar ranking of the importance of factors, then policymakers will be more effective in making policies based on the result of sensitivity analysis methods.

\section{Background}
Sensitivity Analysis is a post-hoc method that has been widely used in the field of deep learning interpretability. It can be categorized into three groups:

\begin{itemize}
\item Back-propagation group: Methods like Deep Lift \cite{shrikumar2019learning} and Integrated Gradients \cite{Sundararajan} use gradient-based approaches to attribute model outputs to input features. 
\item Perturbation group: Methods such as Feature Ablation \cite{meyes2019ablation}and Feature Occlusion \cite{zeiler2014visualizing}  involve altering or masking individual attributes to observe changes in outputs. Morris Method \cite{Morris} involves varying one parameter at a time while holding others constant, often employing a grid of points in the parameter space.
\item Approximation group: Methods such as LIME \cite{ribeiro2016model} and SHAP \cite{lundberg2016consistent} compute an approximate explanation to a complex model, in a simpler and interpretable way such as fitting an Ordinary Least Sqaures (OLS) /Lasso regression.
\end{itemize}
In this work, I will study the perturbation group because of its model-agnostic, non-parametric, and interpretable nature. 

\subsection{Prior Work Done by Applicant}
My research focus is based on two prior works: \textbf{1) Interpreting County-Level COVID-19 Infections Using Deep Learning for Time Series} \cite{10224685}, in which I reproduced existing work, and \textbf{2) 
Interpreting Time Series Transformer Models and Sensitivity Analysis of Population Age Groups to COVID-19 Infections} \cite{islam2023population} project, in which I have conducted sensitivity analysis experiments. We applied the Morris sensitivity analysis method which is defined as: 
\begin{equation} 
\small
\text{Sensitivity}(X, i) = \frac{f(x_1, x_i + \Delta, \ldots, x_k) - f(X)}{\Delta}
\end{equation}
where $\Delta$ is a small change.
In short, the contributions are the following: 
\begin {itemize}
\item 1. Collecting eight different population age group features and COVID-19 cases info for 2years from 3,142 US counties.
\item 2. Train a Temporal Fusion Transformer on the dataset to predict
COVID-19 cases for next 15 days using past 13 days of input.
\item 3. Extend the original Morris method to propose a scaled Morris index. Then calculate the sensitivity of the age group features and rank the age groups by their sensitivity scores. 
\item 4. Use feature ranks to globally interpret the sensitivity of age groups. Then finally evaluate the ranking with ground truth aggregated from reported cases by those age groups.
\end {itemize}

\section{Approach}
Based on the prior work, I will implement several perturbation-based sensitivity analysis methods and apply them to the U.S county-level COVID-19 cases dataset. The features of interest is the percentage of age groups in the population.To examine the robustness of these sensitivity analysis methods, I will train other popular time-series deep learning models like DLinear \cite{dlinear}, Autoformer \cite{autoformer}, PatchTST\cite{patchtst}, and TimesNet\cite{timesnet} to the same U.S county-level COVID-19 cases dataset and apply the sensitivity analysis methods. 

The feature importance produced by the sensitivity analysis methods will be ranked, and we rank the ground truth is aggregated from reported cases by those age groups as well. From here, the evaluation metric, Spearman rank correlation coefficient, can be applied. A value of 1 for the Spearman rank correlation coefficient means perfectly monotonically increasing relationship while a value of -1 implies no monotonic relation.

\section{Evaluation}

\newcolumntype{P}[1]{>{\centering\arraybackslash}p{#1}}
\newcolumntype{M}[1]{>{\centering\arraybackslash}m{#1}}

\begin{table}[htbp]
    \centering
    \small
    \begin{tabular}{|M{2cm}|M{5.5cm}|} \hline
        \textbf{Components}  & \textbf{Implementations}  \\ \hline
    SA Methods & Feature Ablation, Feature Occulsion, and Morris Method \\ \hline
    DL Models & TFT, TimesNet, Autoformer, DLinear, PatchTST\\ \hline
    Dataset & 2-year U.S County-Level COVID cases with 8 age feature groups \\ \hline
    Ground Truth & Cases of certain age groups \\ \hline
    Evaluation & Spearman correlation coefficient \\ \hline
    \end{tabular}
    \caption{Experiment Components and Implementations. ``SA'' refers to Sensitivity Analysis, and ``DL'' refers to Deep Learning}
\end{table}

The experimental setup is described in detail in Table 1. I aim to answer three research questions:
\begin{itemize}
\item Within the same model, do different sensitivity analysis (SA) methods yield comparable outputs and attribute importance rankings? (Validation)
\item Using the same sensitivity analysis method, do different Deep Learning (DL) models impact the output of the sensitivity analysis?  (Robustness) 
\item How well do the results from sensitivity analysis methods align with the ground truth? (Accuracy)
\end{itemize}
With the experimental data, I will first process the data into ranks within each experiment groups and apply Spearman correlation coefficient to observe the consistencies between different groups of experiments. Furthermore, to evaluate which methods produce the best result, their respective Spearman rank correlation coefficient with the ground truth will also be calculated. These experiments aim to provide benchmark results that answer my research questions on DL models and SA methods.

\section{Discussion}
Based on the answers to my three proposed research questions, I can conclude the validation, robustness, and accuracy of sensitivity analysis methods. 
\begin{itemize}
    \item The answer to my first research question will shed light on whether it is necessary to interpret a Deep Learning model with multiple perturbation-group post-hoc method. 
    \item The answer to my second research question will provide insight into the robustness of sensitivity analysis. If they are robust, an reliable interpretation method that is insensitive to different models is achieved. 
    \item The answer to my third research question will inform us on how confident we can trust the results generated from Sensitivity Analysis. 
\end{itemize}
AI transparency can warrent trust, especially for policymakers who are using the outputs from sensitivity analysis to make decisions of great social impact. 

\section{Conclusion}
In conclusion, my research focuses on the validation, robustness, and accuracy of Perturbation-based Sensitivity Analysis Methods. My research aims to evaluate if these methods yield comparable outputs, if different models impact sensitivity analysis results, and how well the feature importance from Sensitivity Analysis Methods align with the ground truth. The results will inform best practices for interpretability in time-series data. Consequently, policymakers can adopt policies with greater confidence, relying on insights from deep learning models and sensitivity analysis. Furthermore, my research carries great value in other related fields. For example, in Uncertainty Quantification, reliable Sensitivity Analysis methods will help determine how much uncertainty an individual source contributes to the total uncertainty in a simulated or experimental quantity, analyzing the effect of noisy data and perturbations. 

\bibliography{aaai24}

\end{document}